\documentclass[sigconf]{acmart}
\AtBeginDocument{%
  }

\copyrightyear{2025}
\acmYear{2025}
\setcopyright{acmlicensed}
\acmConference[MM '25]{Proceedings of the 33rd ACM International Conference on Multimedia}{October 27--31, 2025}{Dublin, Ireland}
\acmBooktitle{Proceedings of the 33rd ACM International Conference on Multimedia (MM '25), October 27--31, 2025, Dublin, Ireland}
\acmDOI{10.1145/3746027.3755114}
\acmISBN{979-8-4007-2035-2/2025/10}

\usepackage{multirow}
\usepackage{enumitem}
\usepackage{makecell}

\begin{document}

\title{HydraMamba: Multi-Head State Space Model for Global Point Cloud Learning}

\author{Kanglin Qu}
\affiliation{%
  \institution{Nanjing University of Aeronautics and Astronautics}
  \city{Nanjing}
  \country{China}
}
\email{klinqu@163.com}
\orcid{0000-0002-5062-5012}

\author{Pan Gao}
\affiliation{%
  \institution{Nanjing University of Aeronautics and Astronautics}
  \city{Nanjing}
  \country{China}
}
\email{Pan.Gao@nuaa.edu.cn}
\orcid{0000-0002-4492-5430}
\authornotemark[1]

\author{Qun Dai}
\affiliation{%
  \institution{Nanjing University of Aeronautics and Astronautics}
  \city{Nanjing}
  \country{China}
}
\email{daiqun@nuaa.edu.cn}
\orcid{0000-0003-4618-7299}
\authornote{Corresponding author.}

\author{Yuanhao Sun}
\affiliation{%
  \institution{Beijing University of Posts and Telecommunications}
  \city{Beijing}
  \country{China}}
\email{sunyh@bupt.edu.cn}
\orcid{0000-0001-9080-5715}


\begin{abstract}
The attention mechanism has become a dominant operator in point cloud learning, but its quadratic complexity leads to limited inter-point interactions, hindering long-range dependency modeling between objects. Due to excellent long-range modeling capability with linear complexity, the selective state space model (S6), as the core of Mamba, has been exploited in point cloud learning for long-range dependency interactions over the entire point cloud. Despite some significant progress, related works still suffer from imperfect point cloud serialization and lack of locality learning. To this end, we explore a state space model-based point cloud network termed HydraMamba to address the above challenges. Specifically, we design a shuffle serialization strategy, making unordered point sets better adapted to the causal nature of S6. Meanwhile, to overcome the deficiency of existing techniques in locality learning, we propose a ConvBiS6 layer, which is capable of capturing local geometries and global context dependencies synergistically. Besides, we propose MHS6 by extending the multi-head design to S6, further enhancing its modeling capability. HydraMamba achieves state-of-the-art results on various tasks at both object-level and scene-level. The code is available at https://github.com/Point-Cloud-Learning/HydraMamba.
\end{abstract}

\ccsdesc[500]{Computing methodologies~Computer vision tasks}

\keywords{3D vision, Point cloud, State space, Locality learning, Serialization}

\maketitle

\section{Introduction}
\label{Section1}

The attention mechanism \cite{1} has been widely used in point cloud learning, and has gradually replaced the convolution and multi-layer perceptron (MLP) as a dominant operator in point cloud learning. Since the quadratic complexity of the attention mechanism causes an unbearable computational overhead when dealing directly with the whole point cloud containing a large number of points, mainstream attention-based networks \cite{2,3,4,5,6,7,8,9,10,11,12,13,14,15,16,17,18,19,20} ingeniously construct local neighborhoods or windows for attention computation. However, this practice inevitably limits the number of points that can be perceived by a single point, hindering long-range dependency modeling between objects.  Hence, exploring networks with both long-range modeling capability and linear complexity is a critical step in the advancement of point cloud learning.

Recently, the state space model (SSM) \cite{21,22,23,24,25,26,27} has achieved impressive results in natural language processing (NLP) by its long-range modeling ability with linear complexity, sparking widespread interest. In particular, the selective state space model (S6) in the related work Mamba \cite{25} achieves flexible selection of relevant information in a data-dependent manner, and this selection mechanism further pushes the long-range modeling capability to a higher level. Meanwhile, S6 is equipped with a hardware-aware algorithm inspired by FlashAttention \cite{28}, significantly improving the running efficiency. These excellent features indicate that S6 has great potential to replace the attention mechanism as the next-generation token mixer in sequence modeling.

Inspired by S6's outstanding performance, some works \cite{58,59,60,61} introduce it into point cloud learning to overcome the trade-off between long-range interactions and computation resources. Despite some significant progress, the above efforts still struggle to yield satisfactory results due to the following two aspects:

\begin{itemize}[leftmargin=0.2cm, rightmargin=0.2cm]
  \item \textbf{Point cloud serialization}. PointMamba \cite{58} recently proposes a serialization based on space-filling curves that establishes more reliable inter-point structure dependencies compared to the axis ordering \cite{59,60}. Specifically, PointMamba concatenates the serialization results of the Hilbert curve and its variant Trans-Hilbert \cite{33}. However, this practice suffers from the following shortcomings: (1) a longer sequence induced by the concatenation introduces redundancy and negatively affects efficiency, (2) concatenating serialized sequences with different spatial relationships tends to cause confusion, and (3) adopting only two serializations may limit capturing spatial information from different perspectives. Consequently, existing efforts have not fully utilized the potential of space-filling curves-based serialization. 
  \item \textbf{Locality learning}. Most works \cite{59,60,61} consider the limitation of S6's unidirectional modeling and achieve a global receptive field on point sequences by introducing the bidirectionality on S6. However, the bidirectional S6 still relies on compressing all contextual information into the history-hidden state for global feature modeling, which leads to inadequate locality learning - a large number of studies have shown \cite{29,30,31,32} that locality learning is crucial for capturing local details such as lines and edges in visual data.
\end{itemize}

Based on the above analyses, we propose a novel point cloud network termed HydraMamba, which obeys the following designs to tackle the above challenges in order to fully unleash the potential of S6 in the point cloud domain:

\begin{itemize}[leftmargin=0.2cm, rightmargin=0.2cm]
  \item \textbf{Shuffle serialization}. We propose a shuffle serialization strategy that enables each layer to infer on structural information from multiple perspectives by randomly assigning six variants of the Hilbert curve to different layers, further enhancing the network's generalization.
  \item \textbf{ConvBiS6 layer}. We propose a ConvBiS6 layer by explicitly adding a branch with 1D grid convolution to the bidirectional S6, and this simple yet effective way compensates for the deficiency of the bidirectional S6 in local geometric modeling, enabling it to extract both local geometries and global context dependencies.
  \item \textbf{Multi-head S6}. Apart from the above designs, inspired by the multi-head self-attention, we also extend the multi-head mechanism to S6 and propose a variant termed MHS6, which has the flexibility to learn different types of temporal dynamics. 
\end{itemize}

To validate the effectiveness of our network, we conduct experiments on various tasks at both object-level and scene-level. Specifically, the object-level tasks include point cloud recognition on the ModelNet40 dataset \cite{35} and part segmentation on the ShapeNet dataset \cite{36}, and the scene-level task contains semantic segmentation on the S3DIS dataset \cite{37}. Notably, with long-range modeling capability with linear complexity, HydraMamba directly performs global interactions over the entire input point cloud, including on scene-level data with more than 100K points, rather than restricting inter-point interactions to the local space due to quadratic complexity as mainstream attention networks do, which fully leverages long-range dependencies between objects, making HydraMamba achieve state-of-the-art results in the above point cloud tasks.

In summary, the contributions of this paper derive from the following three points:

\noindent
(1) A novel network termed HydraMamba is constructed based on the state-space model, which achieves long-range dependency interactions over the entire point cloud through S6's long-range modeling capability with linear complexity.

\noindent
(2) A shuffle serialization strategy and ConvBiS6 layer are designed to help S6 better adapt to point cloud tasks. Specifically, the former constructs multiple structural dependencies in the sequence for S6 to perform geometric inference from different perspectives, while the latter directly exploits the excellent locality-preserving property of the serialization and captures local geometries in the 3D space by 1D grid convolution to overcome the shortcoming of the bidirectional S6 in locality learning.

\noindent
(3) A variant MHS6 on S6 is proposed by introducing the multi-head design, which is capable of flexibly capturing complex spatial dependencies and diverse geometric features from structured point cloud sequences. To our best knowledge, all existing Mamba-based methods are based on S6, and MHS6 is the first exploration of optimizing the computational mode in S6.

\noindent

\section{Related work}
\label{Section2}

\subsection{Attention networks}
\label{Section2.1}

After the great success of Vision Transformer \cite{39} in the image domain, it became a hot topic to study the application of the attention mechanism in point cloud learning. Initially, PCT \cite{40} treats each point as a token and directly performs attention computation on the whole point cloud, but the quadratic complexity of the attention mechanism results in highly computational overhead, making it unable to capture long-range dependencies in large-scale point clouds. Therefore, some works \cite{3,4,8,9,10,18,20} employ the attention mechanism in point neighborhoods constructed for each point through methods such as the K-Nearest neighbor, ball query, and K-Means, but these works are still trapped in high time and space complexity, since a point set has a large cardinality and there is no computational sharing between overlapping neighborhoods. To this end, another class of works \cite{5,6,7,13,14} shifted the attention computation from requiring all points to a finite number of spatial windows, by dividing the 3D space with non-overlapping equal-shaped windows. However, the sparsity of point clouds causes different numbers of points in spatial windows, which introduces the hardware-unfriendly window filling operation to support parallel computation, resulting in inefficiency. Hence, several recent studies \cite{11,12,15} have proposed window partitioning on serialization to construct windows with an equal number of points, where the serialization refers to structuring the point cloud by ordering points according to specific patterns. 
While the above local neighborhoods or windows-based methods advance the attention mechanism in point cloud learning, the neighborhood and window sizes lead to a limited receptive field, hampering long-range dependency interactions between objects.

\begin{figure*}[t]
\centering
\includegraphics[width=5.2in]{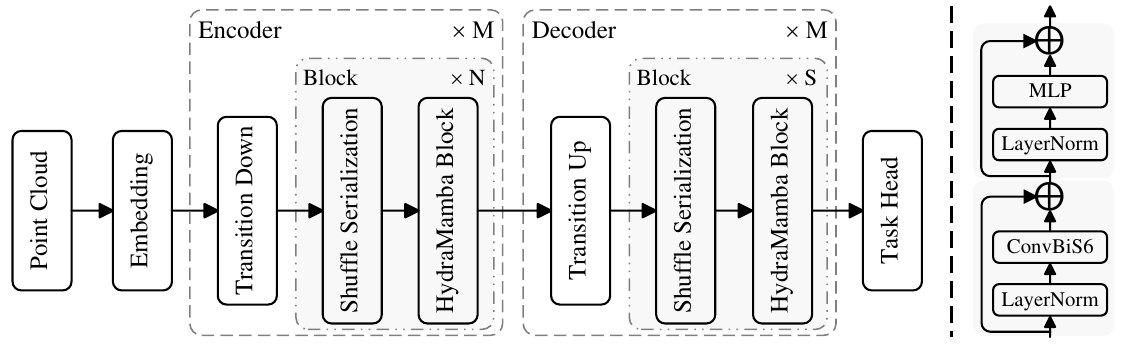}
\caption{Pipeline of HydraMamba (left) and the architecture of HydraMamba block (right).}
\label{Fig1}
\end{figure*}

\subsection{State space models}
\label{Section2.2}

As a mathematical model used in the control theory for describing dynamic systems, the state space model has been introduced to deep learning in recent years to be an effective method for dealing with long sequence modeling tasks. Initially, Gu \textit{et al} proposed a linear state space layer (LSSL) \cite{26} by simulating linear continuous-time state space representations, and demonstrated its superiority in capturing long-range dependencies in the sequence data based on the HiPPO parameter initialization strategy \cite{27}. However, the excessive computational and memory requirements induced by the state representations prevent it from being used as a general-purpose sequence modeling method. To address this bottleneck, the structured state space model (S4) \cite{22} reduces computational resource overhead by normalizing the parameters to a diagonal structure. Subsequently, researchers have explored a variety of SSM variants, including DSS \cite{21}, S5 \cite{23}, and GSS \cite{24}. In particular, S6 is a milestone work as it proposes a selective mechanism that enables the model parameter to vary with the input, which significantly enhances long-range modeling capability through the flexible selection of relevant information. Moreover, it is combined with the simplified H3 architecture \cite{25} to build Mamba, exhibiting excellent performance in NLP. This advancement motivates numerous works to apply S6 to vision tasks, such as video understanding \cite{47,48,49}, medical image segmentation \cite{50,51,52}, and image classification \cite{53,54,55,56,57,89,90,91,92,93}.

Recently, some works have extended the application of S6 to point cloud learning. PointMamba \cite{58} concatenates multiple space-filling curves-based serialization results which produces a longer sequence, resulting in redundancy and inefficiency. Pamba \cite{84} validates its performance only on semantic segmentation tasks. NIMBA \cite{59} and PCM \cite{60} serialize point clouds based on the axis ordering. However, this approach may cause points that are close in the 3D space to not be adjacent in the sequence, generating confusing inter-point structural dependencies. Mamba3D \cite{61} employs S6 for feature channels rather than sequences, and does not take into account the unidirectional modeling property of S6 as our network does. Furthermore, the above works only apply S6 directly to point cloud learning and are mainly evaluated on object-level data, whereas our network not only proposes a stronger variant by introducing the multi-head design, but also demonstrates the potential of S6-based networks on scene-level data.

\section{Preliminaries}
\label{Section3}

\textbf{State space model (SSM)}. The state space model is a time-series modeling framework based on a hidden state $h\left( t \right) \in {{\mathbb{R}}^N}$, which essentially maps an input signal (a one-dimensional function or sequence) $x\left( t \right) \in {{\mathbb{R}}^N}$ to an output signal $y\left( t \right) \in {{\mathbb{R}}^N}$ by the recurrent hidden state evolution. This dynamic process is usually described formally by the following linear ordinary differential equation:
\begin{equation}
h'\left( t \right) = \boldsymbol{A}h\left( t \right) + \boldsymbol{B}x\left( t \right), \;\;\;  y\left(t\right) = \boldsymbol{C}h\left( t \right),
\label{Eq1}
\end{equation}
where $\boldsymbol{A} \in {{\mathbb{R}}^{N \times N}}$ is a state transformation matrix that controls the autonomous evolution of $h\left( t \right)$, $\boldsymbol{B} \in {{\mathbb{R}}^N}$ is an input projection matrix that determines the effect of $x\left( t \right)$ on the state, and $\boldsymbol{C} \in {{\mathbb{R}}^N}$ is an output projection matrix that maps the hidden state to the observation space. It is worth noting that the above mathematical expression of SSM is based on the continuous time domain and is not applicable to discretized sampled signals in deep learning. To solve this problem, the continuous SSM needs to be discretized so that it can handle the input ${x_t}$ on the discrete time step. Among discretization methods, the zero-order hold is a widely adopted scheme due to its clear physical meaning and numerical stability. The discretized equation is as follows:
\begin{equation}
{h_t} = {\boldsymbol{\bar A}}{h_{t - 1}} + {\boldsymbol{\bar B}}{x_t}, \;\;\;  {y_t} = {\boldsymbol{C}}{h_t},
\label{Eq2}
\end{equation}
where the discrete parameters ${\boldsymbol{\bar A}}$ and ${\boldsymbol{\bar B}}$ are derived from the continuous ${\boldsymbol{A}}$ and ${\boldsymbol{B}}$ computed with a sampling interval $\boldsymbol{{\rm{\Delta }}}$, respectively:
\begin{equation}
{\boldsymbol{\bar A}} = {e^{\boldsymbol{{\rm{\Delta }}A}}}, \;\;\;  {\boldsymbol{\bar B}} = {\left( \boldsymbol{{{\rm{\Delta }}A}} \right)^{ - 1}}\left( {{e^{\boldsymbol{{\rm{\Delta }}A}}} - {\boldsymbol{I}}} \right)\left( \boldsymbol{{{\rm{\Delta }}B}} \right).
\label{Eq3}
\end{equation}

\noindent
\textbf{Selective state space model (S6)}. Traditional linear time-invariant state space models use a fixed parameter mechanism, \textit{i.e.}, $\boldsymbol{\bar A}$ and $\boldsymbol{\bar B}$ are invariant, which results in the system adopting the same state evolution rule for all contextual information, thus making it difficult to differentiate between task-relevant crucial information and noise, especially in long-range dependency modeling tasks. Therefore, by integrating an input-driven dynamic parameterization mechanism, the novel S6 is proposed, where $\boldsymbol{\bar A}$ and $\boldsymbol{\bar B}$ are functions of the input. However, the dynamic parameters of S6 prevent it from being computed efficiently using convolution operations as in linear time-invariant SSMs. To this end, S6 introduces a parallel scanning technique with linear complexity \cite{23,62} to ensure efficient computation. Besides, it designs a hardware-aware algorithm through the high-speed SRAM in GPUs, further improving running efficiency.


\section{Methodology}
\label{Section4}

\subsection{Overview}
\label{Section4.1}

As shown in Fig. \ref{Fig1}, HydraMamba follows the standard encoder-decoder architecture in point cloud networks. Initially, a point cloud is transformed into a high-dimensional space through an embedding layer, followed by hierarchical feature aggregation based on the encoder-decoder, and finally an adapted task header is invoked. 
In this study, our network is validated on point cloud segmentation and recognition tasks, where the segmentation header processes the output of the decoder by a multilayer perceptron (MLP) to obtain classification logits for each point, and the recognition header processes the output of the encoder through an average pooling layer and an MLP to obtain global classification logits. Next, we introduce components in the encoder-decoder architecture in detail.

\begin{figure}[t]
\centering
\includegraphics[width=3.in]{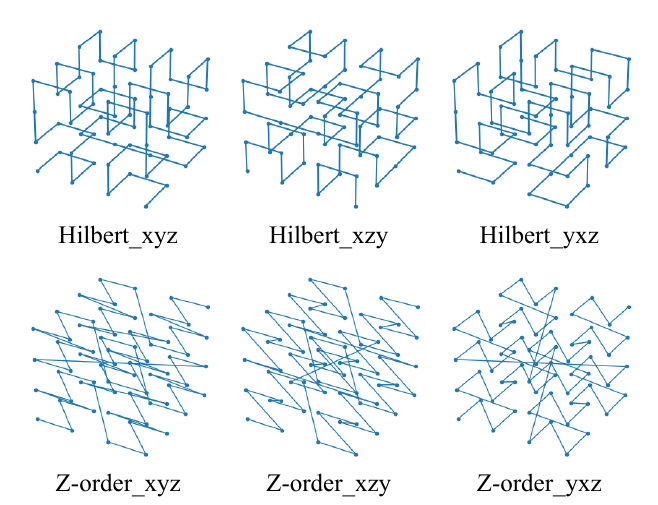}
\caption{Different traversal results of the Hilbert and Z-Order curves in the 3D space, respectively. The suffixes denote the traversal priority, the Hilbert\_xyz, for example, indicates that the traversal is performed with the Hilbert curve in the order of X-axis first, Y-axis second, and Z-axis last.}
\label{Fig2}
\end{figure}

\subsection{Shuffle serialization}
\label{Section4.2}

\textbf{Space-filling curve}. Space-filling curves are a class of paths that cover high-dimensional regions by the continuous single-parameter mapping, whose key property is to transform geometric structures in a high-dimensional space into a one-dimensional sequence while preserving local neighborhood relations. When they are used in point clouds, the high-dimensional space refers to the 3D Euclidean space in which point coordinates are located. Typical space-filling curves include the Hilbert and Z-Order curves, with the Hilbert curve being highly valued for its excellent locality-preserving property and the Z-Order curve known for its high efficiency.

Space-filling curves require that a priority order be specified for traversing a high-dimensional space. As a result, reordering variants of space-filling curves can be obtained by simply altering the priority order of traversal. These variants provide different perspectives for spatial relationships and contexts, compensating for local relationships that may be overlooked in each other. To illustrate this, we present different traversal results for the Hilbert and Z-Order curves in the 3D space, respectively, as shown in Fig. \ref{Fig2} (given that the 3D space contains three axes, space-filling curves are able to produce six different traversals).

OctFormer \cite{12} and Point Transformer V3 \cite{15} are pioneering works that apply space-filling curves to point cloud learning. These studies essentially utilized the spatial proximity of space-filling curves to partition local windows (with arbitrarily varying window shapes) with the same number of points in the 3D space, in order to achieve efficient execution on the hardware. Inspired by this, PointMamba \cite{58} recently uses the spatial proximity of space-filling curves to establish structural dependencies for S6's causal inference. While forming more reliable inter-point structural dependencies compared to the axis ordering \cite{59,60}, it does not fully exploit the potential of space-filling curves-based point cloud serialization, as shown in the analysis in the introduction.

\begin{figure}[t]
\centering
\includegraphics[width=3.3in]{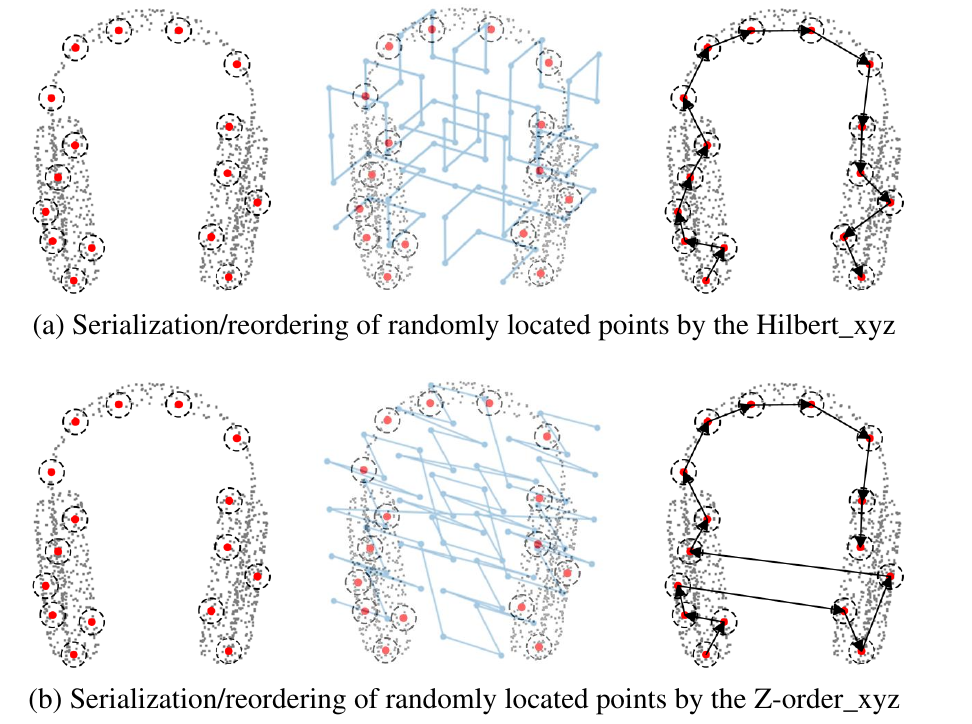}
\caption{Serialization of randomly located points by the Hilbert curve (top) and Z-Order curve (bottom).}
\label{Fig3}
\end{figure}

\begin{figure*}[t]
\centering
\includegraphics[width=4.8in]{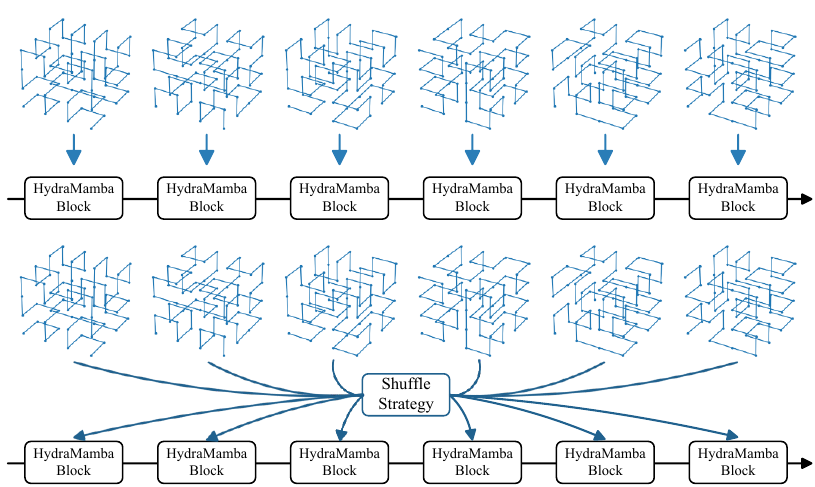}
\caption{Illustrations of the sequential assignment (top) and the shuffle assignment (bottom).}
\label{Fig4}
\end{figure*}

\noindent
\textbf{Shuffle serialization strategy}. Figure \ref{Fig3} shows the serialization of randomly localized points using the Hilbert and Z-Order curves respectively. It is intuitively observed that the Hilbert curve preserves better spatial proximity than the Z-Order curve, which is also confirmed by the previous work \cite{63}. Hence, we establish reliable structural dependencies between points based on the Hilbert curve to achieve accurate causal inference on S6. Considering the aforementioned shortcomings of PointMamba, we sequentially assign the six variants of the Hilbert curve to each HydraMamba block, as shown in Fig. \ref{Fig4} (top), but Tab. \ref{Tab5} reveals that this scheme fails to yield appreciable performance. We attribute this to a single ordering pattern making each HydraMamba block lack the ability to capture geometric relationships from multiple perspectives. For this reason, we propose a shuffle serialization strategy which randomly assigns different patterns of the Hilbert curve to different HydraMamba blocks, as shown in Fig. \ref{Fig4} (bottom), which enables each block to infer spatial relationships from multiple perspectives, further enhancing the network's generalization.

\subsection{HydraMamba block}
\label{Section4.3}

\textbf{Bidirectional S6}. As shown in Section \ref{Section3}, S6 is a forward recurrent process based on the hidden state, \textit{i.e.}, a point in the input sequence can only perceive points indexed before this point, but not interact with points indexed after this point, and such unidirectional modeling is not applicable to visual data requiring global learning. Consequently, we propose a bidirectional S6 to achieve global modeling on the point sequence by introducing the bidirectionality. Specifically, we parallelize two S6s, called Forward S6 and Backward S6 respectively, one performing recurrent scanning from the index direction of the input sequence, while the other from the reverse index direction. In this way, each point in the input sequence possesses a global receptive field.

\begin{figure}[t]
\centering
\includegraphics[width=3.4in]{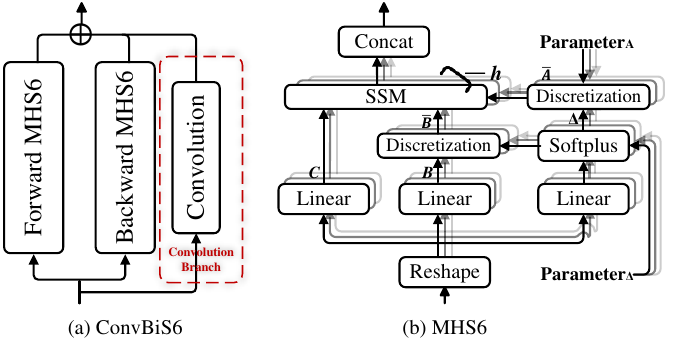}
\caption{Structures of the ConvBiS6 (left) and MHS6 (right).}
\label{Fig5}
\end{figure}

\noindent
\textbf{ConvBiS6}. Since the bidirectional S6 still considers global connectivity by compressing all contextual information into the history-hidden state, it lacks a mechanism for modeling local dependencies between adjacent points. However, locality learning is crucial for visual data because it involves structural details such as lines, edges, and local shapes. Thus, we propose a ConvBiS6 layer based on the bidirectional S6 by explicitly adding a branch with 1D grid convolution, which simultaneously accommodates both global connectivity and local attention, as shown in Fig. \ref{Fig5} (left). The effectiveness of this simple design stems from that the excellent locality-preserving property of the Hilbert curve allows the convolution on the one-dimensional sequence to capture local geometries between points. Furthermore, to enable flexible integration of the ConvBiS6 layer into HydraMamba, it is integrated with the standard Transformer block \cite{1,34} by replacing the attention module, thereby constructing the HydraMamba block, as shown in Fig. \ref{Fig1} (right).

\noindent
\textbf{Multi-head S6}. With dynamic weights, the attention mechanism demonstrates strong global modeling capability, while the multi-head attention endows the attention mechanism with stronger representation capability by decomposing the feature space into multiple heterogeneous subspaces, \textit{i.e.}, each attention head can independently learn interaction patterns at different semantic levels and avoid the overfitting risk of a single attention direction through the parameter difference of the subspace projection matrix. Motivated by this, we propose a corresponding variant termed Multi-head S6 (MHS6) by extending the multi-head design to S6. Algorithm 1 presents the step-wise logical steps of MHS6, (the practical implementation usually processes all the heads in parallel with matrix operations, as shown in Fig. \ref{Fig5} (right)) where the Reshape is an operation for adjusting the data shape, the ${\rm{Linea}}{{\rm{r}}_{N}}$ denotes a linear layer projected to dimension $N$, and the ${\rm{Concat}}$ represents a concatenation operation at the channel level. Each head in MHS6 learns meaningful time steps and different types of temporal dynamics, 
and this extreme flexibility not only enhances the model's ability to infer complex geometric relationships, but also improves robustness by the multi-feature fusion. Hence, MHS6 replaces S6 in the ConvBiS6 layer as the token mixer in the HydraMamba block.

\subsection{Transition down and transition up}
\label{Section4.4}

The transition down is to reduce the cardinality of a point set layer-wise in the encoder to achieve passing geometric features down hierarchically. Mainstream transition down modules include the farthest point sampling (FPS) \cite{65} and grid pooling \cite{4}. The former maximizes information utilization while decreasing point cloud resolution by uniformly covering and preserving key structures. Due to iteratively selecting points furthest from the selected point set, this method is time-consuming and cannot be applied to large-scale point clouds. The latter significantly optimizes the processing efficiency by dividing points into grid spaces and aggregating features, but it may omit key points owing to local focusing compared to the global uniform coverage of FPS.

The transition up is to recover the cardinality of a point set in the decoder according to the corresponding encoder stage and pass structural semantics up hierarchically. Popular transition up modules include the linear interpolation \cite{65} and grid unpooling \cite{4}. The former interpolates directly using the spatial distribution of adjacent points, which better preserves the local continuity of point clouds, but when it is applied to large-scale point clouds, determining adjacent points for each point results in inefficiency. The latter, paired with the grid pooling, is an efficient method that directly maps features on a point to points in the grid space corresponding to the grid pooling process.

In summary, HydraMamba applies the FPS and linear interpolation when we deal with object-level tasks, and for scene-level tasks we employ the grid pooling and grid unpooling.

\begin{table}[t]
  \centering
\renewcommand{\arraystretch}{1}
\resizebox{\linewidth}{!}{
  \begin{tabular}{l}
    \toprule
    \textbf{Algorithm 1} Multi-head S6 (MHS6)  \\
    \midrule
    \textbf{Input}: $\boldsymbol{x}$: $\left( {B,\;L,\;D} \right)$ and and the number of heads $h$  \\
    \textbf{Output}: $\boldsymbol{y}$: $\left( {B,\;L,\;D} \right)$  \\
    1. $\boldsymbol{x}$: ($B, L, h, D/h$) $ \leftarrow $ ${\rm{Reshape}}\left( {\boldsymbol{x},\;\left( {B,\;L,\;h,\;D/h} \right)} \right)$  \\
    2. \textbf{for} each head $i \in \left[ {0,\;h - 1} \right]$:  \\
    3. $\;\;\;$ $\boldsymbol{B}$: $\left( {B,\;L,\;N} \right)$ $ \leftarrow $ ${\rm{Linea}}{{\rm{r}}_{N}}\left( {\boldsymbol{x}\left[ {:,\;:,\;i,\;:} \right]} \right)$  \\
    4. $\;\;\;$ $\boldsymbol{C}$: $\left( {B,\;L,\;N} \right)$ $ \leftarrow $ ${\rm{Linea}}{{\rm{r}}_{N}}\left( {\boldsymbol{x}\left[ {:,\;:,\;i,\;:} \right]} \right)$  \\
    /* shape of ${\bf{Paramete}}{{\bf{r}}_{\bf{\Delta }}}$ is ($D/h$), and the softplus ensures  \\   
     positive $\boldsymbol{{\rm{\Delta }}}$ */  \\
    5. $\;\;\;$ $\boldsymbol{\rm{\Delta }}$: ($B, L, D/h$) $ \leftarrow $ $\log \left( {1 + {e^{\left( {{\rm{Linea}}{{\rm{r}}_{D/h}}\left( {\boldsymbol{x}\left[ {:,\;:,\;i,\;:} \right]} \right) + {\bf{Paramete}}{{\bf{r}}_{\boldsymbol{\rm{\Delta }}}}} \right)}}} \right)$  \\
    /* shape of ${\bf{Paramete}}{{\bf{r}}_{\boldsymbol{A}}}$ is ($D/h, N$). Each dimension in \\ 
     $\boldsymbol{A}$ represents a structured diagonal ${N} \times {N}$ matrix \cite{22}  */ \\
    6. $\;\;\;$ ${\boldsymbol{\bar A}}$: ($B, L, D/h, N$) $ \leftarrow $ ${\boldsymbol{\rm{\Delta }}}  {\boldsymbol{\otimes}}  {\bf{Paramete}}{{\bf{r}}_{\boldsymbol{A}}}$  \\
    7. $\;\;\;$ ${\boldsymbol{\bar B}}$: ($B, L, D/h, N$) $ \leftarrow $ ${\boldsymbol{\rm{\Delta }}}  {\boldsymbol{\otimes}}  \boldsymbol{B}$  \\
    /* time-variant: calculated by the parallel scanning */  \\
    8. $\;\;\;$ ${{\boldsymbol{y}}^i}$: ($B, L, D/h$) $ \leftarrow $ SSM$\left( {{\boldsymbol{\bar A}},\;{\boldsymbol{\bar B}},\;{\boldsymbol{C}}} \right)\left( {{\boldsymbol{x}}\left[ {:,\;:,\;i,\;:} \right]} \right)$  \\
    9. \textbf{end}  \\
    10. ${\boldsymbol{y}} = {\rm{Concat}}\left( {{{\boldsymbol{y}}^0},\;{{\boldsymbol{y}}^1},\; \cdots ,\;{{\boldsymbol{y}}^{h - 1}}} \right)$  \\
    return $\boldsymbol{y}$  \\
    \bottomrule
  \end{tabular}}
\end{table}

\section{Experiment}
\label{Section5}

To validate that our network, based on S6, is able to fully utilize long-range dependencies in the entire point cloud 
, we compare it with different types of networks on multiple tasks at object-level and scene-level. Moreover, we analyze the efficiency of our network by comparing it with previous state-of-the-art works on memory overhead and inference latency. In addition, based on extensive ablation studies, we explore the effectiveness of each key design.

\begin{table}[t]
  \centering
\renewcommand{\arraystretch}{0.8}
\setlength{\tabcolsep}{1.5mm}{
\resizebox{\linewidth}{!}{
  \begin{tabular}{cccc}
    \toprule
Networks &	Reference & Backbone & OA (\%)  \\
\midrule
PointNet++ \cite{65} & NIPS 2017 & MLP & 91.9  \\
PointNext \cite{66} & NIPS 2022 & MLP & 93.2  \\
PointMLP \cite{67} & ICLR 2022 & MLP & 93.6  \\
\midrule
PointWeb \cite{69} & CVPR 2019 & CNN & 92.3  \\
AdaptConv \cite{71} & ICCV 2021 & CNN & 93.4  \\
3D-GCN \cite{72} & PAMI 2022 & CNN & 92.1  \\
\midrule
PointASNL \cite{19} & CVPR 2020 & Attention & 93.2  \\
PCT \cite{40} & CVM 2021 & Attention & 93.2  \\
Point Transformer \cite{3} & ICCV 2021 & Attention & 93.7  \\
Point-BERT \cite{18} & CVPR 2022 & Attention & 93.2  \\
Point-MAE \cite{20} & ECCV 2022 & Attention & 93.8  \\
OctFormer \cite{12} & TOG 2023 & Attention & 92.7  \\
ACT \cite{17} & ICLR 2023 & Attention & 93.5  \\
PointConT \cite{16} & JAS 2024 & Attention & 93.5  \\
PointGST \cite{87} & arXiv 2024 & Attention & 93.4 \\
LCM \cite{88} & NIPS 2024 & Attention & 93.6  \\
\midrule
PointMamba \cite{58} & NIPS 2024 & SSM & 92.4  \\
PCM \cite{60} & arXiv 2024 & SSM & 92.6  \\
PointTramba \cite{74} & arXiv 2024 & SSM \& Attention & 92.7  \\
OctMamba \cite{75} & arXiv 2024 & SSM & 92.7  \\
Mamba3D \cite{61} & MM 2024 & SSM & 93.4  \\
HydraMamba & - & SSM & \textbf{94.0}  \\
    \bottomrule
  \end{tabular}}}
  \caption{Recognition results of different backbones-based networks on the ModelNet40 dataset.}
  \label{Tab1}
\end{table}

\subsection{Comparison on different tasks}
\label{Section5.1}

\textbf{Object-level recognition}. The ModelNet40 is a popular object-level point cloud recognition dataset, containing 12,311 CAD models with 40 categories, of which 9,843 objects are used for training and the remaining for testing. Following the data preprocessing procedure of Qi \textit{et al} \cite{65}, we uniformly sample 1024 points and their corresponding normal vectors from each sample as the input. As per most relevant studies in the literature, the overall accuracy (OA) is used as an evaluation metric. Tab.\ref{Tab1} lists the experimental results of our work and different backbones-based networks on the ModelNet40 dataset. By comparing with the attention networks, we observe that the existing SSM networks yield only suboptimal results and do not fully utilize the long-range modeling capability of the state space model. However, our work addresses their shortcomings and achieves state-of-the-art performance by the related designs, achieving 94.0\% OA and outperforming all the existing architectures with different paradigms. This breakthrough indicates that SSM is able to serve as a competitive alternative to mainstream attention-based frameworks in point cloud learning. 



\noindent
\textbf{Object-level part segmentation}. The ShapeNet is an object-level point cloud part segmentation dataset, containing 16,878 samples with 50 parts from 16 categories, of which 14,005 objects are used for training and the remaining for testing. The same data preprocessing procedure is followed as for the ModelNet40 dataset, and the instance mIoU (Ins. mIoU) is used as an evaluation metric. Tab. \ref{Tab2} lists the experimental results of our work and different backbones-based networks on the ShapeNet dataset. HydraMamba achieves state-of-the-art performance in point cloud part segmentation with 86.8\% Ins. mIoU, which outperforms all the existing methods in mainstream architectures. Notably, it sets a new performance ceiling for SSM-based networks, which marks the first time that the SSM architecture outperforms the Transformer-based paradigm in point cloud understanding, challenging the dominance of the attention mechanism in this domain. 


\begin{table}[t]
  \centering
\renewcommand{\arraystretch}{0.8}
\setlength{\tabcolsep}{1.mm}{
\resizebox{\linewidth}{!}{
  \begin{tabular}{cccc}
    \toprule
Networks &	Reference & Backbone & Ins. mIoU (\%)  \\
\midrule
PointNet \cite{64} & CVPR 2017 & MLP & 83.7  \\
PointNet++ \cite{65} & NIPS 2017 & MLP & 85.1  \\
PointMLP \cite{67} & ICLR 2022 & MLP & 86.1  \\
ReCon \cite{76} & ICML 2023 & MLP & 86.4  \\
\midrule
SpiderCNN \cite{68} & ECCV 2018 & CNN & 85.3  \\
DGCNN \cite{70} & TOG 2019 & CNN & 85.2  \\
AdaptConv \cite{71} & ICCV 2021 & CNN & 86.4  \\
\midrule
PCT \cite{40} & CVM 2021 & Attention & 86.4  \\
Point Transformer \cite{3} & ICCV 2021 & Attention & 86.6  \\
Point-BERT \cite{18} & CVPR 2022 & Attention & 85.6  \\
MaskPoint \cite{73} & ECCV 2022 & Attention & 86.0  \\
Point-MAE \cite{20} & ECCV 2022 & Attention & 86.1  \\
Patchformer \cite{8} & CVPR 2022 & Attention & 86.5  \\
IDPT \cite{85} & ICCV 2023 & Attention & 85.7 \\
APES \cite{77} & CVPR 2023 & Attention & 85.8  \\
ACT \cite{17} & ICLR 2023 & Attention & 86.1  \\
DAPT \cite{86} & CVPR 2024 & Attention & 85.5 \\
\midrule
PCM \cite{60} & arXiv 2024 & SSM & 84.3  \\
Mamba3D \cite{61} & MM 2024 & SSM & 85.6  \\
PointTramba \cite{74} & arXiv 2024 & SSM \& Attention & 85.7  \\
PointMamba \cite{58} & NIPS 2024 & SSM & 85.8  \\
HydraMamba & - & SSM & \textbf{86.8}  \\
    \bottomrule
  \end{tabular}}}
  \caption{Part segmentation results of different backbones-based networks on the ShapeNet dataset.}
  \label{Tab2}
\end{table}

\noindent
\textbf{Scene-level semantic segmentation}. The S3DIS is a scene-level semantic segmentation dataset, including 3D scanned points from 271 rooms in 6 regions, with 13 semantic labels. Unlike the previous attention networks \cite{3,4,40,79,83}, which are limited by quadratic complexity and divide the whole scene into blocks to be processed separately, HydraMamba takes directly the entire point cloud as the network input, based on its long-range modeling capability with linear complexity, in order to take full advantage of long-range dependencies between objects. As per most relevant studies in the literature, we use the region 5 as test set and the other regions for training, and adopt the mean IoU (mIoU) as an evaluation metric. Tab. \ref{Tab3} lists the experimental results of our work and different backbones-based networks on the S3DIS dataset. HydraMamba achieves an excellent result of 73.6\% mIoU on the S3DIS dataset, surpassing all the existing methods. Notably, with its long-range modeling with linear complexity, HydraMamba takes directly the entire scene point cloud in the S3DIS as the network input, which contains more than 100K points. Furthermore, HydraMamba demonstrates our network's excellent long range modeling capability by surpassing Point Transformer V3.





\subsection{Efficiency analysis}
\label{Section5.2}

To further analyze the efficiency of HydraMamba, we compare it with several previous state-of-the-art methods on the S3DIS dataset, and the latency and memory footprint in a single inference are used as evaluation metrics. Specifically, to ensure a fair comparison, the latency and memory footprint in a single inference are taken as the average values obtained over the entire S3DIS test set with a batch size of 1 on the same RTX 4090 GPU. Tab. \ref{Tab4} lists the latency and memory footprint of our work and multiple previous state-of-the-art methods in a single inference. Benefiting from S6's linear memory overhead and the hardware-aware algorithm, our network achieves a breakthrough in memory efficiency, with only 5.9G memory footprint, exceeding the optimal Point Transformer V3. In terms of the latency, HydraMamba achieves a speedup of 25.4\% and 2.8x over OctFormer and Point Transformer V2, respectively, but is still slightly slower than Point Transformer V3. We attribute this to the following two aspects: (1) Point Transformer V3 divides points in the entire scene into windows with an equal number of points, which can be processed in parallel by the self-attention mechanism at the block level, whereas HydraMamba processes the entire scene at once in a global receptive field, where each point is computed sequentially. However, this computational mode of our network is a key to enabling long-range dependency modeling on the entire point cloud, which we argue is beneficial overall, despite inevitably leading to inefficiency. (2) Transformer-based networks benefit from optimization algorithms dedicated to the attention mechanism in the hardware \cite{28}, whereas SSM, as an emerging model, has not been more fully explored in terms of acceleration compared to the well-studied attention mechanism. Overall, the low memory and high efficiency properties of our network reflect its unique value in resource-constrained scenarios.

\begin{table}[t]
  \centering
\renewcommand{\arraystretch}{0.8}
\setlength{\tabcolsep}{1.8mm}{
\resizebox{\linewidth}{!}{
  \begin{tabular}{cccc}
    \toprule
Networks &	Reference & Backbone & mIoU (\%)  \\
\midrule
RandLA-Net \cite{80} & PAMI 2022 & MLP & 63.2  \\
RepSurf \cite{81} & CVPR 2022 & MLP & 68.9  \\
PointNext \cite{66} & NIPS 2022 & MLP & 70.5  \\
PointMeta \cite{82} & CVPR 2023 & MLP & 69.5  \\
\midrule
DGCNN \cite{70} & TOG 2019 & CNN & 56.5  \\
KPConv \cite{32} & ICCV 2019 & CNN & 67.1  \\
3D-GCN \cite{72} & PAMI 2022 & CNN & 58.6  \\
\midrule
PCT \cite{40} & CVM 2021 & Attention & 61.3  \\
Point Transformer \cite{3} & ICCV 2021 & Attention & 70.4  \\
Point Transformer V2 \cite{4} & NIPS 2022 &  Attention & 71.6  \\
SuperpointTransformer \cite{83} & ICCV 2023 & Attention & 68.9  \\
Siwn3D \cite{79} & arXiv 2023 & Attention & 72.5  \\
Point Transformer V3 \cite{15} & CVPR 2024 & Attention & 73.4  \\
\midrule
Pamba \cite{84} & AAAI 2025 & SSM & 73.5  \\
PCM \cite{60} & arXiv 2024 & SSM & 63.4  \\
HydraMamba & - & SSM & \textbf{73.6}  \\
    \bottomrule
  \end{tabular}}}
  \caption{Semantic segmentation results of different backbones-based networks on the S3DIS dataset.}
  \label{Tab3}
\end{table}

\subsection{Ablation studies}
\label{Section5.3}

To fully unlock the potential of S6 in the point cloud domain, HydraMamba proposes several key designs, including the shuffle serialization strategy, ConvBiS6, and MHS6. Next, we validate the effectiveness of these designs by ablation comparison.

\noindent
\textbf{Shuffle serialization strategy}. The shuffle serialization strategy enables each block to infer spatial relationships from multiple perspectives by introducing randomness and diversity. We validate the effectiveness of our strategy by comparing it with sequentially assigning the six variants and randomly assigning fewer variants, respectively. Tab. \ref{Tab6} lists the experimental results of different settings-based serialization strategies, and reveals three key conclusions: first, the shuffle mechanism exhibits a significant advantage over the fixed sequential assignment (Group I vs Group II, 93.96\% vs 93.23\%), suggesting that the perspective diversity introduced by randomness breaks the limitations of patterned feature learning. Second, the number of the variants is strongly positively correlated with classification performance. With the number of the variants gradually increasing from 2 to 6 (Group VIII$ \rightarrow $I), the OA presents an increasing trend, which proves that multiple Hilbert variants effectively enhance the representation ability of the model to complex geometric structures. Finally, the serialization strategy without the Hilbert curve (Group IX) is significantly worse than the other settings, which both indicates that S6's causal nature requires causal dependencies between tokens, and underlines that our strategy is capable of efficiently constructing structural dependencies between points for causal inference, resolving the conflict between the causal nature of S6 and the disorder of point sets.




\begin{table}[t]
  \centering
\renewcommand{\arraystretch}{0.88}
\setlength{\tabcolsep}{1.2mm}{
\resizebox{\linewidth}{!}{
  \begin{tabular}{cccc}
    \toprule
Networks & Reference & Latency (ms) & Memory (G)  \\
\midrule
Siwn3D \cite{79} & arXiv 2023 & 365 & 10.7  \\
Point Transformer V2 \cite{4} & NIPS 2022 & 153 & 22.2  \\
OctFormer \cite{12} & TOG 2023 & 71 & 15.2  \\
Point Transformer V3 \cite{15} & CVPR 2024 & \textbf{49} & 6.3  \\
HydraMamba & - & 54 & \textbf{5.9}  \\
    \bottomrule
  \end{tabular}}}
  \caption{Latency and memory footprint of our network and multiple previous state-of-the-art methods in a single inference on the S3DIS dataset.}
  \label{Tab4}
\end{table}

\begin{table}[t]
  \centering
\renewcommand{\arraystretch}{0.89}
\setlength{\tabcolsep}{1.4mm}{
\resizebox{\linewidth}{!}{
  \begin{tabular}{ccccccccc}
    \toprule
\multirow{2}{*}{Group} & \multirow{2}{*}{Random} & \multicolumn{6}{c}{Six variants} & \multirow{2}{*}{OA}  \\ \cline{3-8}
 & & xyz & yxz & xzy & zxy & yzx & zyx &  \\
\midrule
I & \checkmark & \checkmark & \checkmark & \checkmark & \checkmark & \checkmark & \checkmark & \textbf{93.96}  \\
II & $ \times $ & \checkmark & \checkmark & \checkmark & \checkmark & \checkmark & \checkmark & 93.23  \\
III & \checkmark & \checkmark & \checkmark & \checkmark & \checkmark & $ \times $ & $ \times $ & 93.19  \\
IV & \checkmark & \checkmark & \checkmark & $ \times $ & $ \times $ & \checkmark & \checkmark & 93.15  \\
V & \checkmark & $ \times $ & $ \times $ & \checkmark &	\checkmark & \checkmark & \checkmark & 93.11  \\
VI & \checkmark & \checkmark & \checkmark & $ \times $ & $ \times $ & $ \times $ & $ \times $ & 93.11  \\
VII & \checkmark & $ \times $ & $ \times $ & \checkmark & \checkmark & $ \times $ & $ \times $ & 92.79  \\
VIII & \checkmark & $ \times $ & $ \times $ & $ \times $ & $ \times $ & \checkmark & \checkmark & 92.75  \\
IX & $ \times $ & $ \times $ & $ \times $ & $ \times $ & $ \times $ & $ \times $ & $ \times $ & 89.38  \\
    \bottomrule
  \end{tabular}}}
  \caption{Experimental results of different settings-based serialization strategies on the ModelNet40 dataset.}
  \label{Tab5}
\end{table}

\noindent
\textbf{ConvBiS6}. The ConvBiS6 layer introduces bidirectionality and locality by the bidirectional S6 and convolution branch, respectively, enabling the network to synergistically capture local geometries and global contextual dependencies. To validate its effectiveness, we compare the experimental results with and without the bidirectional S6 and convolution branch. Besides, we perform experiments on the convolution branch using either the traditional convolution or depth-wise convolution \cite{38} to verify the effect of different types of convolution operators on the ConvBiS6. Tab. \ref{Tab6} lists the experimental results of different settings-based ConvBiS6. By observing Groups III and V, it is found that the network using only the bidirectional S6 achieves an absolute performance gain of 3.27\%, validating the decisive impact of the bidirectional S6 through global context modeling. Additionally, using the convolution branch alone only improves 0.75\%, while adding the convolution branch on the bidirectional S6 achieves a higher performance gain, suggesting that local feature extraction needs to be combined with global dependency capture for better performance. Notably, as shown in Groups I and II, the OA difference between the depth-wise convolution and traditional convolution is only 0.01\%, indicating that the specific type of the convolution operator has a negligible impact on the performance, while locality itself is more critical. 


\noindent
\textbf{MHS6}. By introducing the multi-head design on S6, MHS6 is able to flexibly capture complex spatial dependencies from the structured point sequence. To validate its effectiveness, we compare the experimental results with and without the multi-head mechanism on S6. In particular, our code sets the initial number of heads corresponding to the initial embedding dimension to 6 by default, and the number of heads changes proportionally with the embedding dimension. Hence, we further explore the effect of the head size on the network by setting different initial head sizes. Tab. \ref{Tab7} reveals two key findings: first, the multi-head structure significantly improves the network performance, as compared between Group I and Groups II-IV, indicating that parallelized multi-subspace feature interactions effectively enhance local geometric perception. Second, the number of heads shows a nonlinear correlation with the performance, when the initial number of heads is increased from 3 to 6, the OA improves by 0.52\% to reach a peak of 93.96\%, indicating that a smaller number of heads cannot adequately learn meaningful time steps and different types of temporal dynamics, whereas a further increase in the number of heads to 12 results in a performance drop of 0.44\%, suggesting that an excessive number of heads triggers the degradation of low-dimensional subspaces. The above reveals that it is necessary to adapt the number of heads to the embedding dimension.



\begin{table}[t]
  \centering
\renewcommand{\arraystretch}{0.88}
\setlength{\tabcolsep}{2.0mm}{
\resizebox{\linewidth}{!}{
  \begin{tabular}{ccccc}
    \toprule
\multirow{2}{*}{Group} & \multirow{2}{*}{Bidirectionality} & \multicolumn{2}{c}{Convolution branch} & \multirow{2}{*}{OA}  \\ \cline{3-4}
 & & Depth-wise & Traditional &  \\
\midrule
I & \checkmark & \checkmark & $ \times $ & \textbf{93.96}  \\
II & \checkmark & $ \times $ & \checkmark & 93.95  \\
III & \checkmark & $ \times $ & $ \times $ & 92.40  \\
IV & $ \times $ & \checkmark & $ \times $ & 89.88  \\
V & $ \times $ & $ \times $ & $ \times $ & 89.13  \\
    \bottomrule
  \end{tabular}}}
  \caption{Experimental results of different settings-based ConvBiS6 on the ModelNet40 dataset.}
  \label{Tab6}
\end{table}

\begin{table}[t]
  \centering
\renewcommand{\arraystretch}{0.86}
\setlength{\tabcolsep}{5mm}{
\resizebox{\linewidth}{!}{
  \begin{tabular}{cccc}
    \toprule
Group & MHS6 & Number of heads & OA  \\
\midrule
I & $ \times $ & - & 93.13  \\
II & \checkmark & 3 & 93.44  \\
III & \checkmark & 6 & \textbf{93.96}  \\
IV & \checkmark & 9 & 93.66  \\
V & \checkmark & 12 & 93.52  \\
    \bottomrule
  \end{tabular}}}
  \caption{Experimental results of different settings-based MHS6 on the ModelNet40 dataset.}
  \label{Tab7}
\end{table}

\section{Conclusion}
\label{Section6}

In this paper, we propose a novel state space model-based network termed HydraMamba, which overcomes the shortcomings of existing works that introduce S6 into point cloud learning, including imperfect point cloud serialization and lack of locality learning. Specifically, we design a shuffle serialization strategy by introducing randomness and diversity to demonstrate the potential of space-filling curves in point cloud serialization.
Moreover, we propose a ConvBiS6 layer which is capable of synergistically capturing both local geometries and global contextual dependencies. Besides, we ingeniously extend the multi-head design in the multi-head attention mechanism to S6 and propose a variant called MHS6, pushing S6's long-range modeling capability to a higher level.

With its excellent long-range modeling capability with linear complexity, HydraMamba takes full advantage of long-range dependencies between objects in the entire point cloud and achieves superior performance on various tasks at object-level and scene-level. However, even with linear complexity, it still does not outperform Point Transformer V3, a state-of-the-art work in well-studied attention networks, in the inference latency. As a result, it is a future endeavor to fully investigate the state space model in terms of acceleration and push it forward towards efficiency.

\section*{Acknowledgments}

This work is supported in part by the National Natural Science Foundation of China under Grant (62476126, 62272227).

\bibliographystyle{ACM-Reference-Format}
\bibliography{sample-base}

\newpage
\appendix

\section{Visualizations}

In order to intuitively demonstrate the part segmentation performance of our network on the ShapeNet dataset, we list in Fig. \ref{Fig6} the visualization results of our network, PointMamba \cite{58}, previously the best performing in SSM-based networks, and Point Transformer \cite{3}, the best performing in attention-based networks, where the red points denote that these points are misclassified. The comparison of the visualization results reveals that our network is able to achieve better part segmentation results at the boundaries of objects. Moreover, we visualize the semantic segmentation results of our network for different scenarios in the S3DIS dataset, as shown in Fig. \ref{Fig7}.

\begin{figure*}[h]
\centering
\includegraphics[width=5.0in]{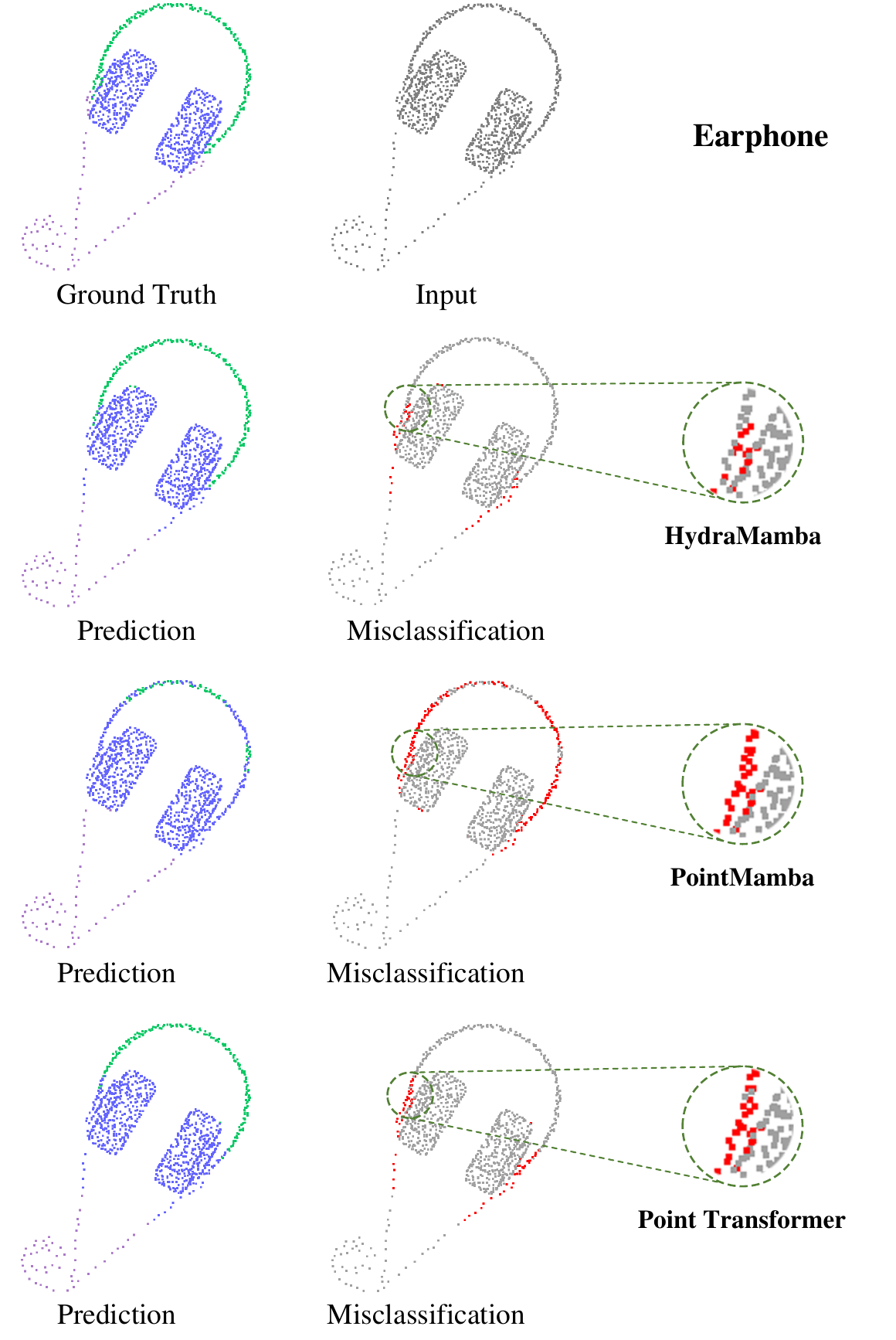}
\end{figure*}

\begin{figure*}[h]
\centering
\includegraphics[width=5.0in]{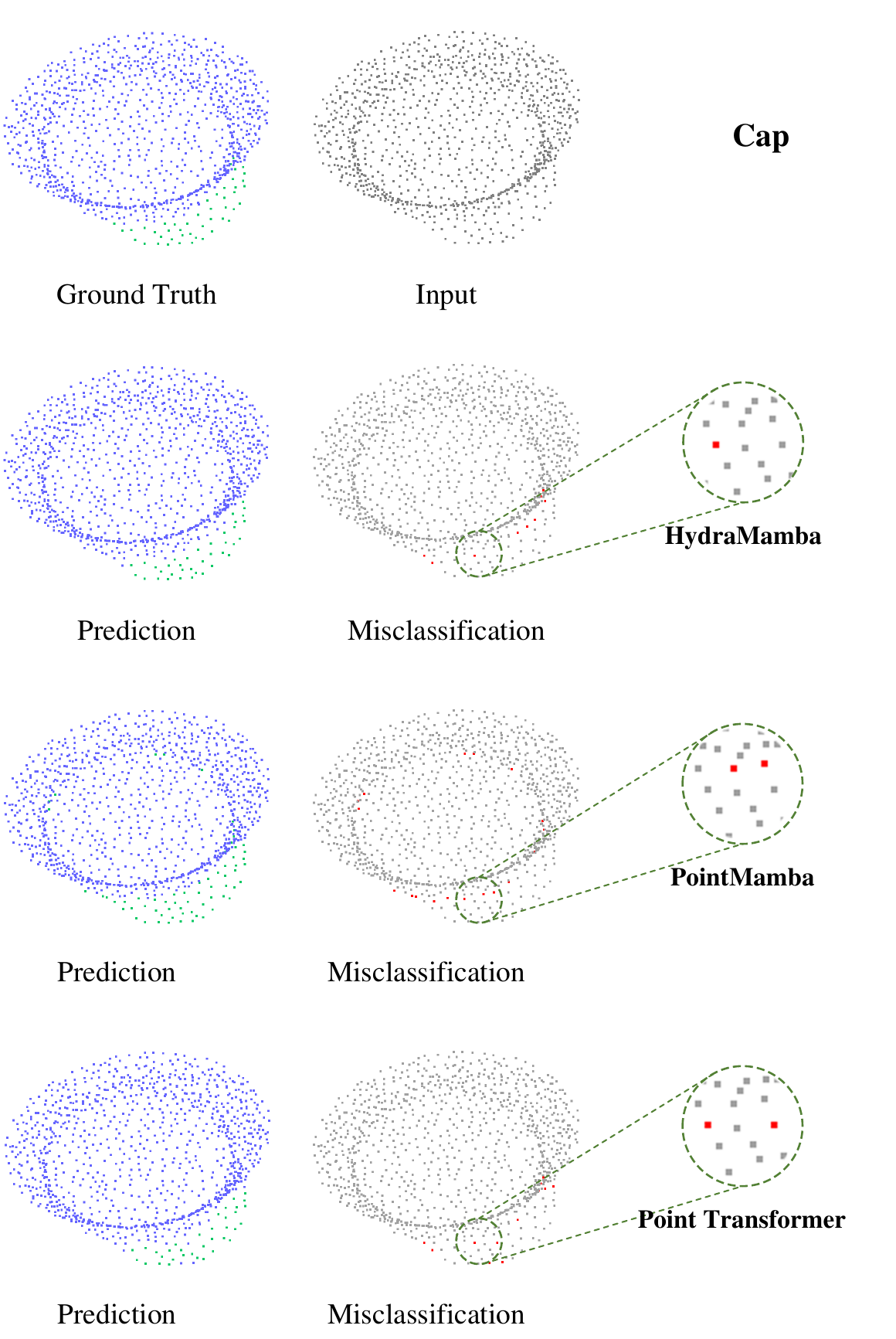}
\end{figure*}

\begin{figure*}[h]
\centering
\includegraphics[width=5.0in]{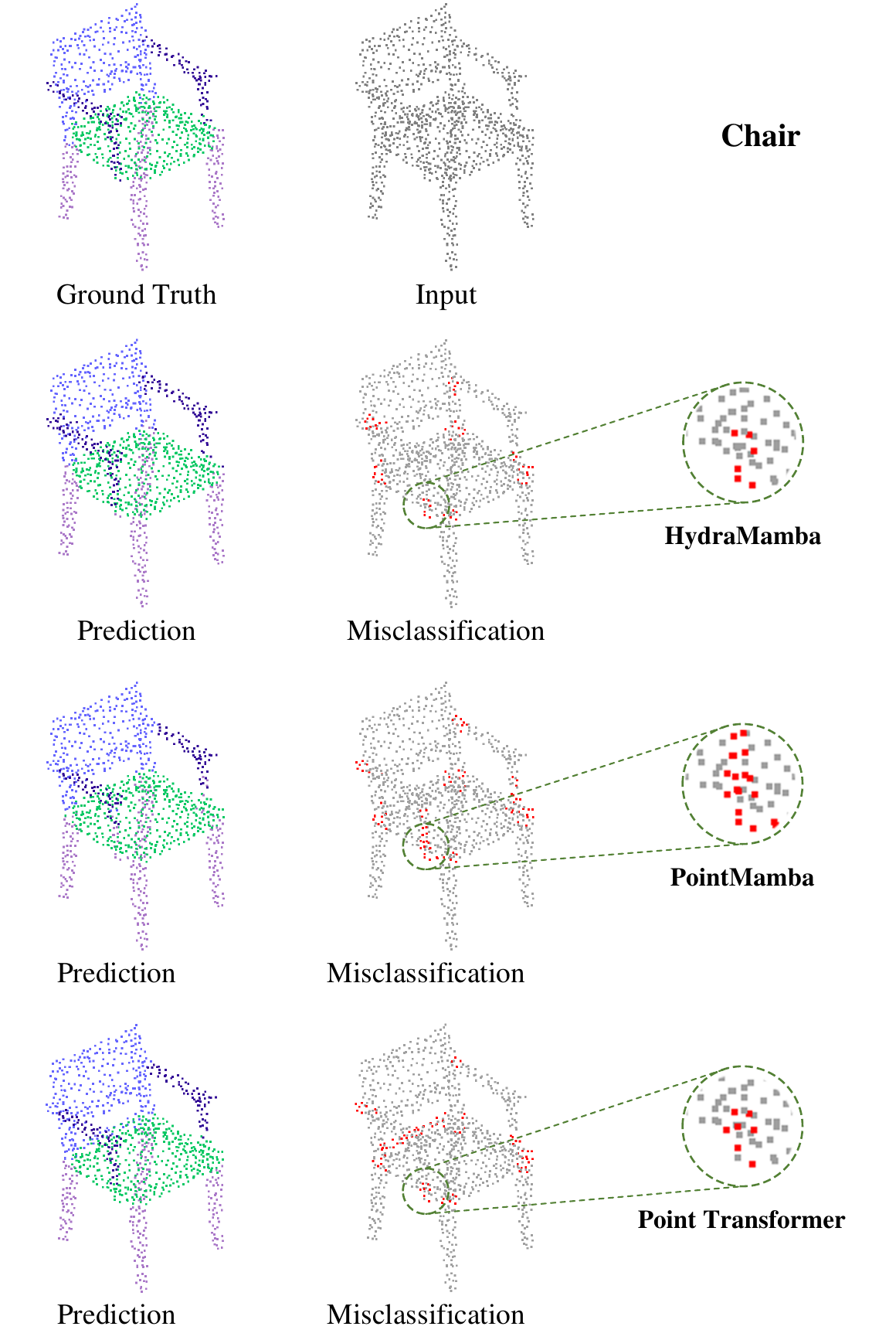}
\end{figure*}

\begin{figure*}[h]
\centering
\includegraphics[width=5.0in]{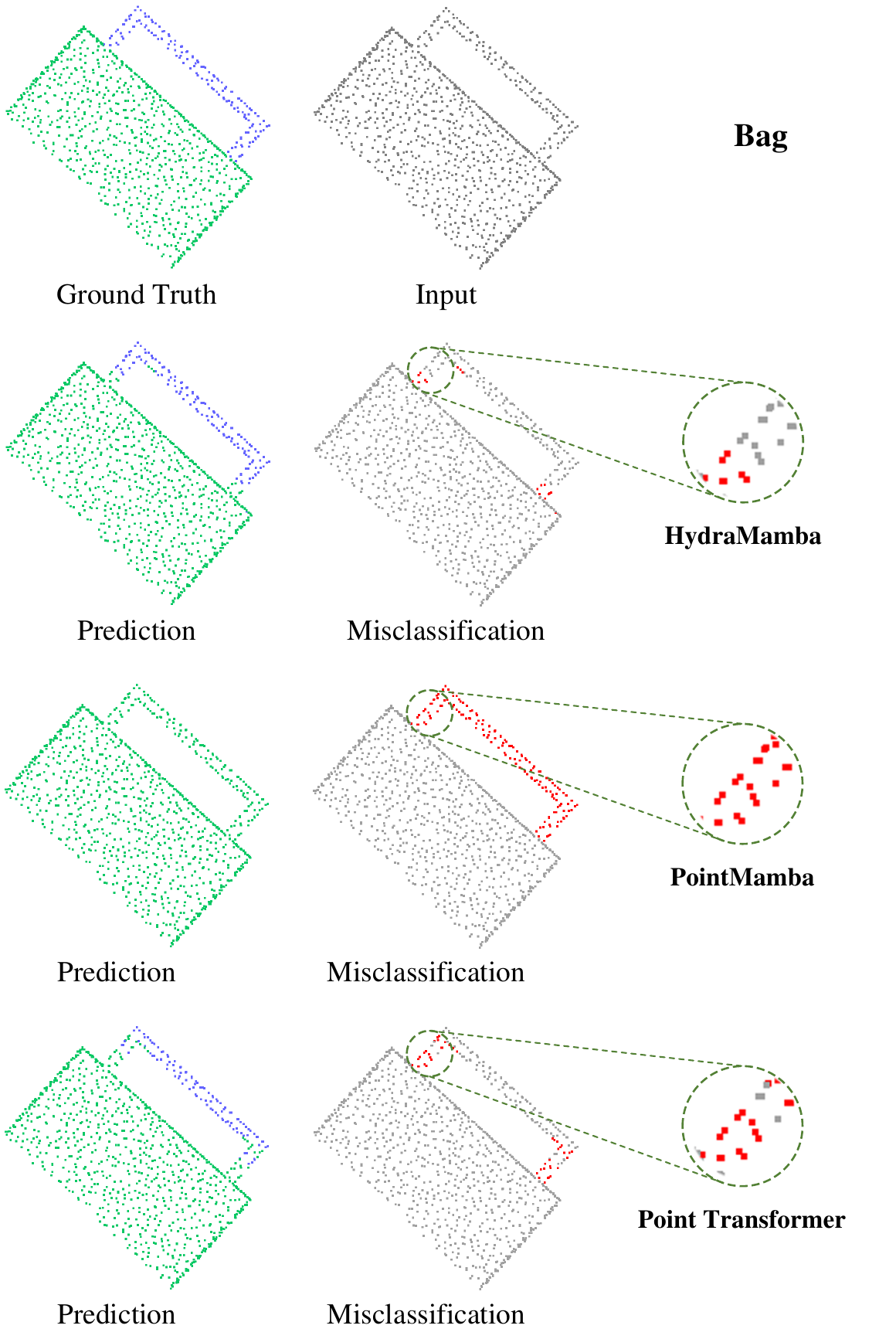}
\end{figure*}

\begin{figure*}[h]
\centering
\includegraphics[width=5.0in]{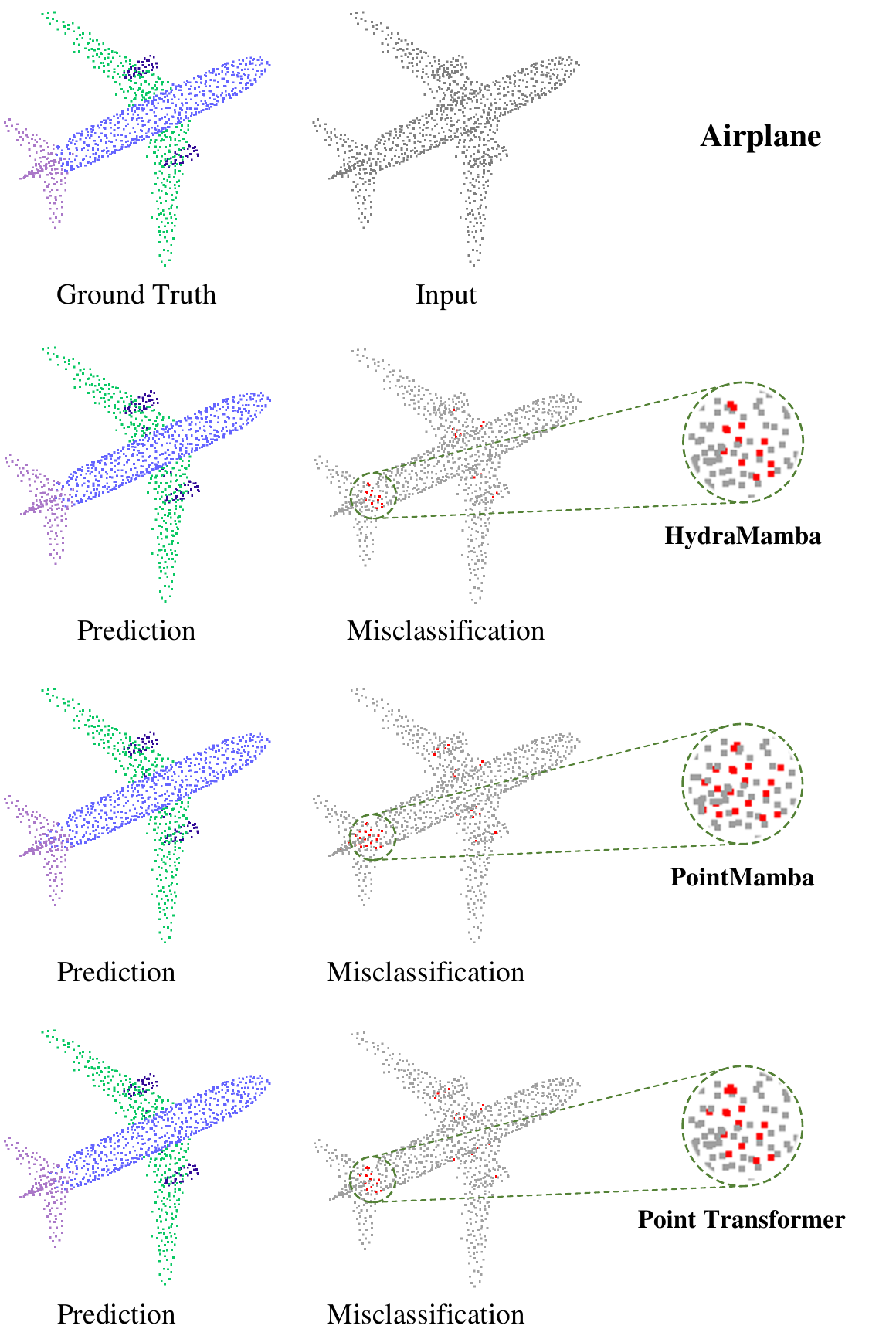}
\end{figure*}

\begin{figure*}[h]
\centering
\includegraphics[width=5.0in]{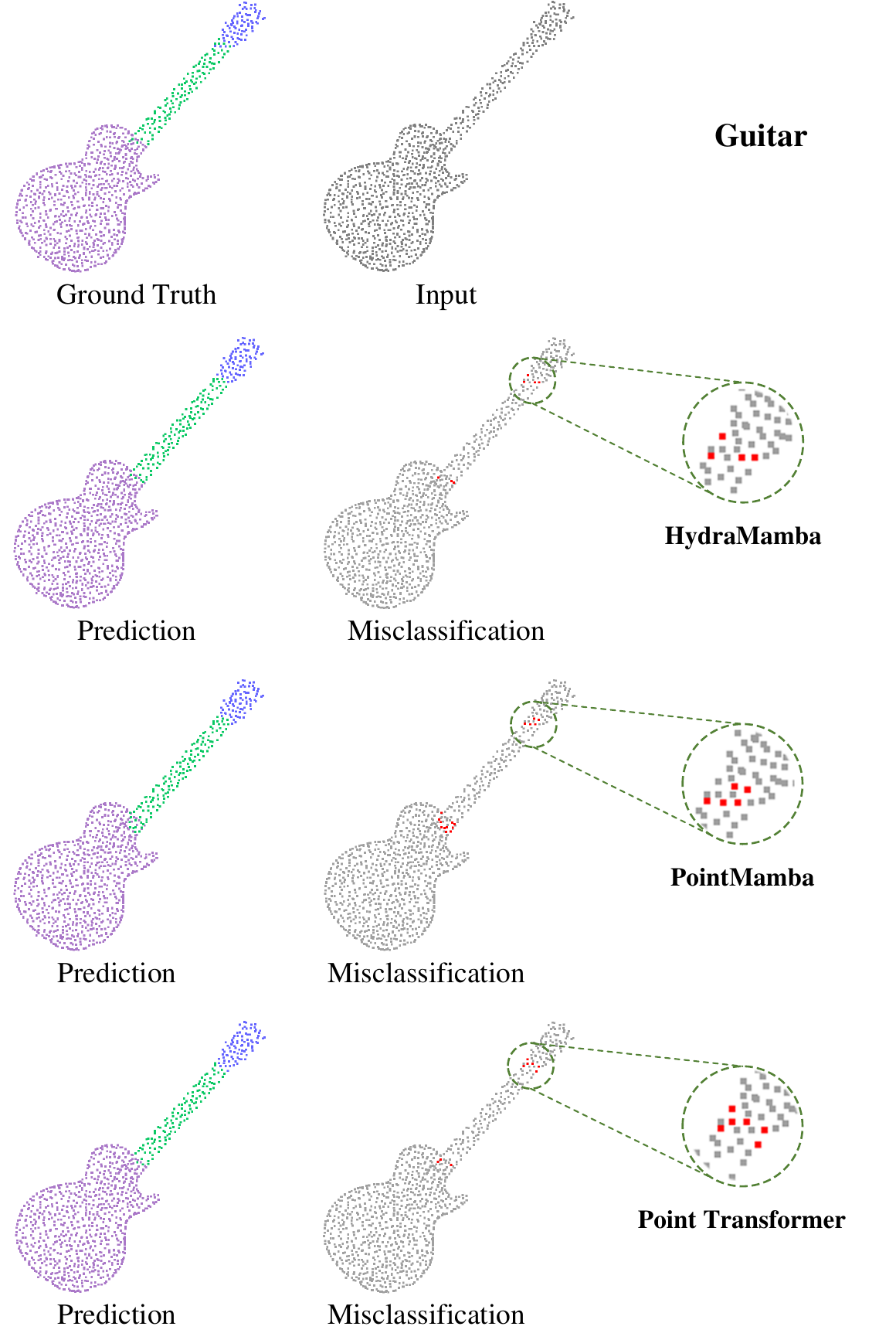}
\caption{Visualization results of HydraMamba, PointMamba, and Point Transformer in the ShapeNet dataset.}
\label{Fig6}
\end{figure*}

\begin{figure*}[h]
\centering
\includegraphics[width=6.2in]{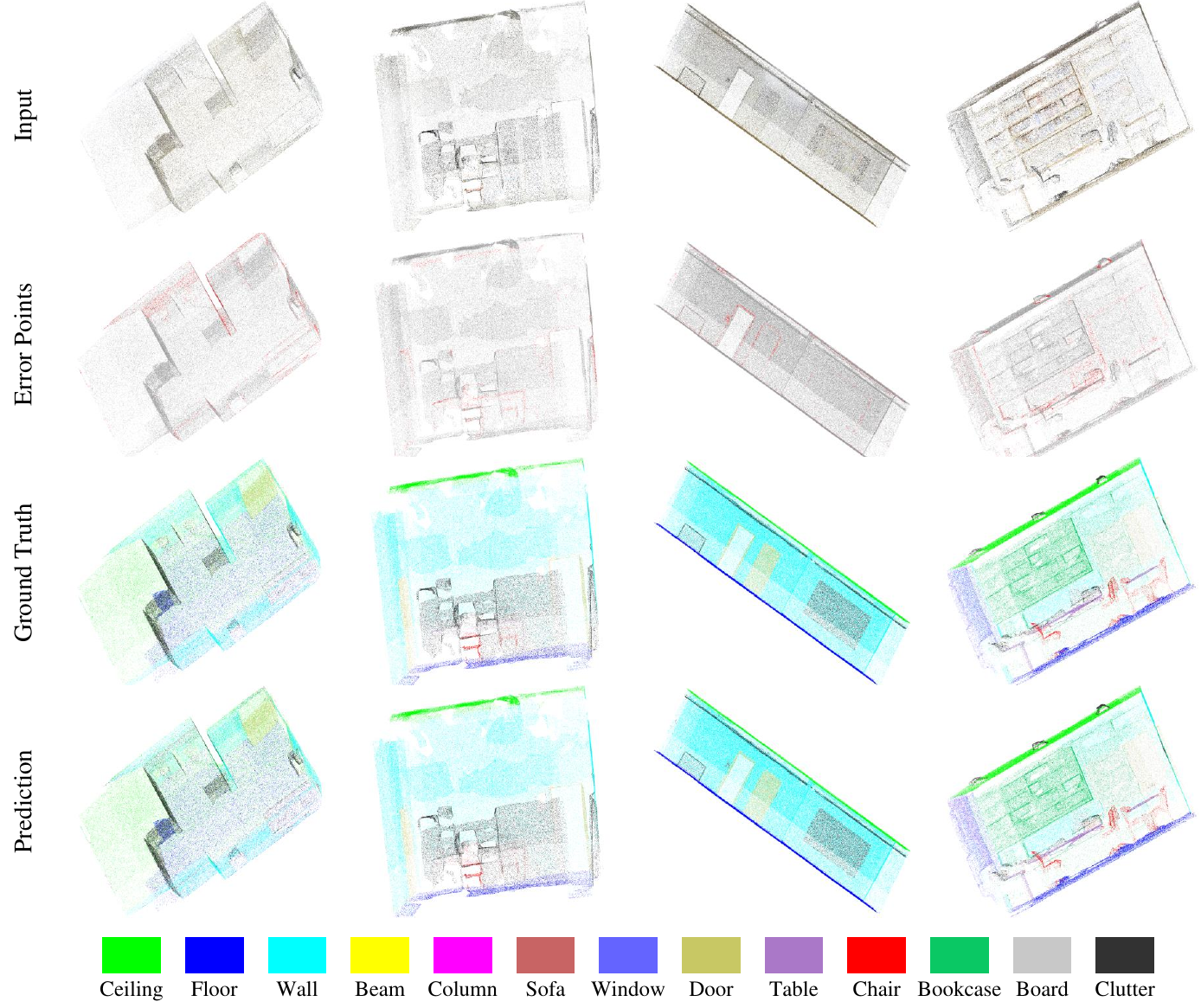}
\caption{Visualization results of HydraMamba on different scenarios in the S3DIS dataset.}
\label{Fig7}
\end{figure*}

%
%
%
%
%

\section{More experimental results}

To further validate the robustness of the network to complex scenarios, the sixth column in Tab. \ref{Tab8} lists the experimental results of our network and different operators-based works on challenging ScanObjectNN (PB\_T50\_RS) dataset, where CloudMamba outperforms the state-of-the-art PointConT and PCM, which strongly proves its robustness. 

\begin{table}[h]
  \centering
\renewcommand{\arraystretch}{0.8}
\setlength{\tabcolsep}{3.0mm}{
\resizebox{\linewidth}{!}{
  \begin{tabular}{cccc}
    \toprule
Networks &	Reference & Backbone & OA (\%)  \\
\midrule
PointNet++ \cite{65} & NIPS 2017 & MLP & 77.9  \\
PointNext \cite{66} & NIPS 2022 & MLP & 87.7  \\
PointMLP \cite{67} & ICLR 2022 & MLP & 85.4  \\
\midrule
DGCNN \cite{70} & TOG 2019 & CNN & 78.1  \\
\midrule
Point-BERT \cite{18} & CVPR 2022 & Attention & 83.1  \\
Point-MAE \cite{20} & ECCV 2022 & Attention & 85.2  \\
ACT \cite{17} & ICLR 2023 & Attention & 87.9  \\
PointConT \cite{16} & JAS 2024 & Attention & 88.0  \\
PointGST \cite{87} & arXiv 2024 & Attention & 85.6 \\
LCM \cite{88} & NIPS 2024 & Attention & 87.8  \\
\midrule
PointMamba \cite{58} & NIPS 2024 & SSM & 87.3  \\
PCM \cite{60} & arXiv 2024 & SSM & 88.1  \\
Mamba3D \cite{61} & MM 2024 & SSM & 87.6  \\
HydraMamba & - & SSM & \textbf{88.3}  \\
    \bottomrule
  \end{tabular}}}
  \caption{Recognition results of different backbones-based networks on the ScanObjectNN dataset.}
  \label{Tab8}
\end{table}

\section{Connections with image tasks}

Recently, there has been a rapid development in applying Mamba to image tasks. Related researches have been conducted from two aspects: (1) how to adapt images to the causal nature of Mamba and (2) how to make the unidirectional modeling of Mamba satisfy the requirement of global context for image understanding. The former has been extensively studied and many approaches have been proposed, such as scanning the entire row or column axis \cite{54,53,89,50}, scanning in a skipping step \cite{90}, scanning the row or column axis in local windows \cite{55}, scanning in a tree topology \cite{56,57,91}, and one recent study has employed a deformable scanning \cite{92}. The above methods are based on the grid structure property of images, and when these methods are applied to point clouds they cause the collapse of geometric structures due to the irregularity and disorder of point clouds. Therefore, we utilize space-filling curves to establish structural dependencies between points, which fully takes into account point clouds' unique data characteristics different from those of images, thereby adapting our network to the causal nature of Mamba. For the latter, the majority of approaches solve it by the bidirectional S6, but they tend to neglect the importance of locality learning. The recent work SegMAN \cite{93} overcomes this shortcoming by incorporating neighborhood attention on the sliding-window of images. However, the sliding-window of point clouds requires a neighborhood query for each point by using KNN or ball query, which will lead to large time and computation overheads. In contrast, our convolution branch directly exploits the excellent locality-preserving property of the serialization from the previous stage to capture local geometries between points in 3D space, which is more efficient and beneficial for scalability to large-scale point clouds. It is worth noting that Mamba-based methods in the image field are all based on S6, whereas our work devises a stronger variant of S6 called MHS6, which is aligned with the multi-head attention mechanism, enabling image tasks to benefit from its flexibility in learning different types of temporal dynamics. 

\end{document}